\documentclass[letterpaper, 10 pt, conference]{ieeeconf}  % Comment this line out if you need a4paper

\IEEEoverridecommandlockouts                              % This command is only needed if 
\usepackage[table,dvipsnames,svgnames,x11names]{xcolor}
\definecolor{yellow}{rgb}{1, 1, 0.7}
\definecolor{orange}{rgb}{1, 0.85, 0.7}
\definecolor{tablered}{rgb}{1, 0.7, 0.7}
\definecolor{red}{rgb}{1, 0, 0}

\usepackage{cite}
\usepackage{fancyhdr, graphics, graphicx} % for pdf, bitmapped graphics files
\usepackage{epsfig} % for postscript graphics files
\usepackage{mathptmx} % assumes new font selection scheme installed
\usepackage{times} % assumes new font selection scheme installed
\usepackage{amsmath} % assumes amsmath package installed
\usepackage{amssymb}  % assumes amsmath package installed
\usepackage{amsfonts}
\usepackage[ruled,vlined]{algorithm2e}
\usepackage{algorithmic}
\usepackage{bm}
\usepackage{booktabs}
\usepackage{dcolumn,tipa}
\newcolumntype{d}[1]{D{.}{.}{#1}}
\usepackage{etoolbox}
\AtBeginEnvironment{tcolorbox}{\small}
\usepackage{fixltx2e}
\usepackage{gensymb}
\usepackage{listings}
\usepackage{makecell}
\usepackage{multirow}
\usepackage{pifont}
\usepackage[most]{tcolorbox}
\usepackage{textcomp}
\usepackage{comment}
\usepackage{todonotes}
\usepackage{hyperref}
\usepackage{cleveref}
\usepackage{caption}
\usepackage{subcaption}

\usepackage{xspace}
\newcommand{\etal}{\textit{et~al}.\xspace}
\tcbuselibrary{breakable}
\usepackage[activate={true,nocompatibility},final,tracking=true,kerning=true,spacing=true,factor=1100,stretch=10,shrink=10]{microtype}
\microtypesetup{protrusion=false}
\newcommand*\colorcheck[1]{%
  \expandafter\newcommand\csname #1check\endcsname{\textcolor{#1}{\ding{51}}}%
}
\newcommand*\colorcross[1]{%
  \expandafter\newcommand\csname #1cross\endcsname{\textcolor{#1}{\ding{55}}}%
}
\colorcheck{green}
\colorcross{red}
\newcommand{\boldparagraph}[1]{\vspace{0.5em}\noindent{\bf #1.}}
\renewcommand{\baselinestretch}{0.982}
\let\origmaketitle\maketitle

\makeatletter
\newcommand{\smallerfootnotes}{%
  \let\orig@makefntext\@makefntext
  \renewcommand\@makefntext[1]{%
    \parindent 1em%
    \noindent
    \tiny
    \hb@xt@1.8em{\hss\@makefnmark}##1}
}

\renewcommand\maketitle{%
  \let\@makefntext\orig@makefntext
  \origmaketitle
  \smallerfootnotes
}
\makeatother

\smallerfootnotes

\title{\LARGE \bf
Context-Aware Human Behavior Prediction Using Multimodal Large Language Models: Challenges and Insights
}

\author{Yuchen Liu$^{1,2}$, Lino Lerch$^{3}$, Luigi Palmieri$^{1}$, Andrey Rudenko$^{1}$, Sebastian Koch$^{1,3}$,\\Timo Ropinski$^{3}$, Marco Aiello$^{2}$% <-this % stops a space
\thanks{$^{1}$Corporate Sector Research and Advance Engineering, Robert Bosch GmbH, Germany, {\tt\small \{yuchen.liu2, luigi.palmieri, andrey.rudenko, sebastian.koch2\}@de.bosch.com}}%
\thanks{$^{2}$Service Computing Department, Institute of Architecture of Application Systems, University of Stuttgart, Germany, {\tt\small marco.aiello@iaas.uni-stuttgart.de}} 
\thanks{$^{3}$Visual Computing Group, Ulm University, Germany, {\tt\small \{lino.lerch, sebastian.koch, timo.ropinski\}@uni-ulm.de}} %
\thanks{This work was partly supported by the EU Horizon 2020 research and innovation program under grant agreement No. 101017274 (DARKO).}
}

\begin{document}
\bstctlcite{IEEEexample:BSTcontrol}

\maketitle

\begin{abstract}
Predicting human behavior in shared environments is crucial for safe and efficient human-robot interaction. Traditional data-driven methods to that end are pre-trained on domain-specific datasets, activity types, and prediction horizons.
In contrast, the recent breakthroughs in Large Language Models (LLMs) promise open-ended cross-domain generalization to describe various human activities and make predictions in any context.
In particular, Multimodal LLMs (MLLMs) are able to integrate information from various sources, achieving more contextual awareness and improved scene understanding.
The difficulty in applying general-purpose MLLMs directly for prediction stems from their limited capacity for processing large input sequences, sensitivity to prompt design, and expensive fine-tuning. 
In this paper, we present a systematic analysis of applying pre-trained MLLMs for context-aware human behavior prediction. To this end, we introduce a modular multimodal human activity prediction framework that allows us to benchmark various MLLMs, input variations, In-Context Learning (ICL), and autoregressive techniques. Our evaluation indicates that the best-performing framework configuration is able to reach 92.8\% semantic similarity and 66.1\% exact label accuracy in predicting human behaviors in the target frame.
Project webpage: \url{https://cap-mllm.github.io/}
\end{abstract}

\section{Introduction}
\label{sec:intro}
\begin{figure}
    \centering
    \includegraphics[width=1\linewidth]{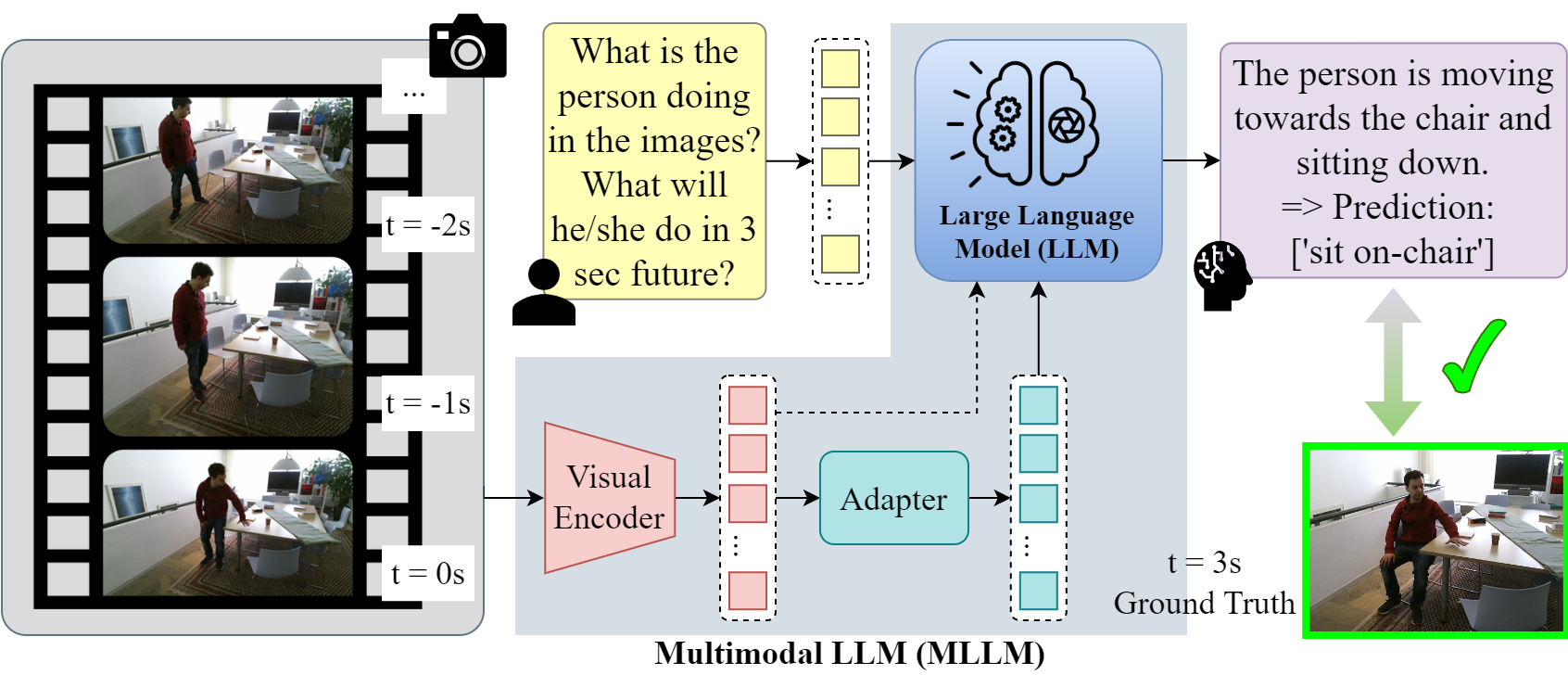}
    \caption{Context-aware human behavior prediction using Multimodal Large Language Model (MLLM). The MLLM (shaded zone) is the main component of the prediction system. It consists of an LLM, a visual encoder, and an adapter (e.g., MLP, Q-Transformer, or cross-attention layers for generating tokens passed to the LLM). The MLLM takes the user instruction and the historical visual observation as input, and forecasts the human behavior.\vspace{-1.5em}}
    \label{fig:ArchBehPred}
\end{figure}

The ability to accurately predict human behavior is essential for robotic systems operating in environments shared with humans \cite{rudenko2020human}. This skill is critical for ensuring safety and efficiency across a wide range of applications, including autonomous driving \cite{zhang2023pedestrian}, intralogistics \cite{jahanmahin2022human}, and home automation \cite{cao_long-term_2020}. Predictive collision avoidance \cite{stefanini2024ral}, anticipatory human-robot interaction (HRI) \cite{lasota2019robust} and safe navigation \cite{kruse2013human} all rely on the system’s capacity to anticipate human actions and movements effectively.
Traditional learning-based methods struggle with transferring to unseen environments and activities that were not part of the training set \cite{li2021toward}. 
In contrast, Large Language Models (LLMs) exhibit promising generalization capabilities. Leveraging rich contextual embeddings and vast commonsense knowledge, LLMs can handle a large variety of environments and domains,
considerably improving the performance and efficiency of the downstream tasks \cite{liu2024delta}.
However, directly applying LLMs for human behavior prediction is challenging since they can only process unimodal information, i.e., textual tokens, and lack visual understanding capabilities. As a result, they often hallucinate the contents of the scene, which is particularly problematic for long-term, real-world human behavior forecasting tasks \cite{zhao2024wildhallucinations}.
Extended with visual encoders and adapters, Multimodal Large Language Models (MLLMs) can process several types of inputs and ground them in the real world, as illustrated in Fig.~\ref{fig:ArchBehPred}. This grounding makes MLLMs suitable for a wide range of vision-enabled robotics applications and offers promising avenues for predicting human behavior by combining visual scene understanding with natural language reasoning \cite{chen2024motionllm}.

With the goal of analyzing the applicability of pre-trained MLLMs for context-aware human behavior prediction, we introduce a modular MLLM-based prediction framework and conduct rigorous experiments analyzing the impact of different model components on prediction performance. 
In particular, we make use of several representative proprietary and open-source general-purpose MLLM backbones, In-Context-Learning (ICL) prompt techniques, and encodings of historical observations described with past activity labels, images, and scene captioning. We also review the effect of varying the numbers of ICL examples (including zero-shot) and applying autoregressive vs.\ direct output. 

Prior art human behavior prediction systems typically address the problem of predicting the end-point of the video sequence as early as possible, using the few initial frames as input. These methods are pre-trained on a given domain and produce a single action label as the prediction.
In contrast, we propose to use general-purpose MLLMs to predict potentially several future activities at an arbitrary future time instance.
To the best of our knowledge, this is the first attempt to systematically analyze how MLLMs can be applied to anticipate potentially multiple human activities in the same frame. To that end, we formulate and evaluate several hypotheses on building a performant prediction system:

\textbf{H1:} MLLMs are capable of correctly predicting non-trivial sequences of human activities (i.e., such where the target label is different from the latest observation, or where multiple correct activity labels are given as the ground truth).

\textbf{H2:} Adding visual context yields more accurate predictions, compared to using purely text-based LLMs.

\textbf{H3:} Prediction accuracy generally improves with the number of ICL examples, compared to zero-shot prediction.

\textbf{H4:} Predicting intermediate actions between the current and the target time frame in an autoregressive manner improves the accuracy.

In our extensive evaluation on PROX and PROX-S datasets \cite{hassan_resolving_2019, zhao_compositional_2022}, the best-performing configuration (in terms of the visual representation type, number of ICL examples, and autoregressive prediction method) is able to reach $92.8\%$ semantic similarity and $66.1\%$ exact label accuracy in predicting multiple human behaviors in the target frame.
Our findings indicate that a small number of ICL examples benefits the performance of the state-of-the-art MLLMs.
Compared to textual inputs, images provide significant additional value in improving visual scene understanding capability, and image captioning further improves prediction accuracy.

\begin{figure*}[th!]
    \centering
    \includegraphics[width=0.9\textwidth]{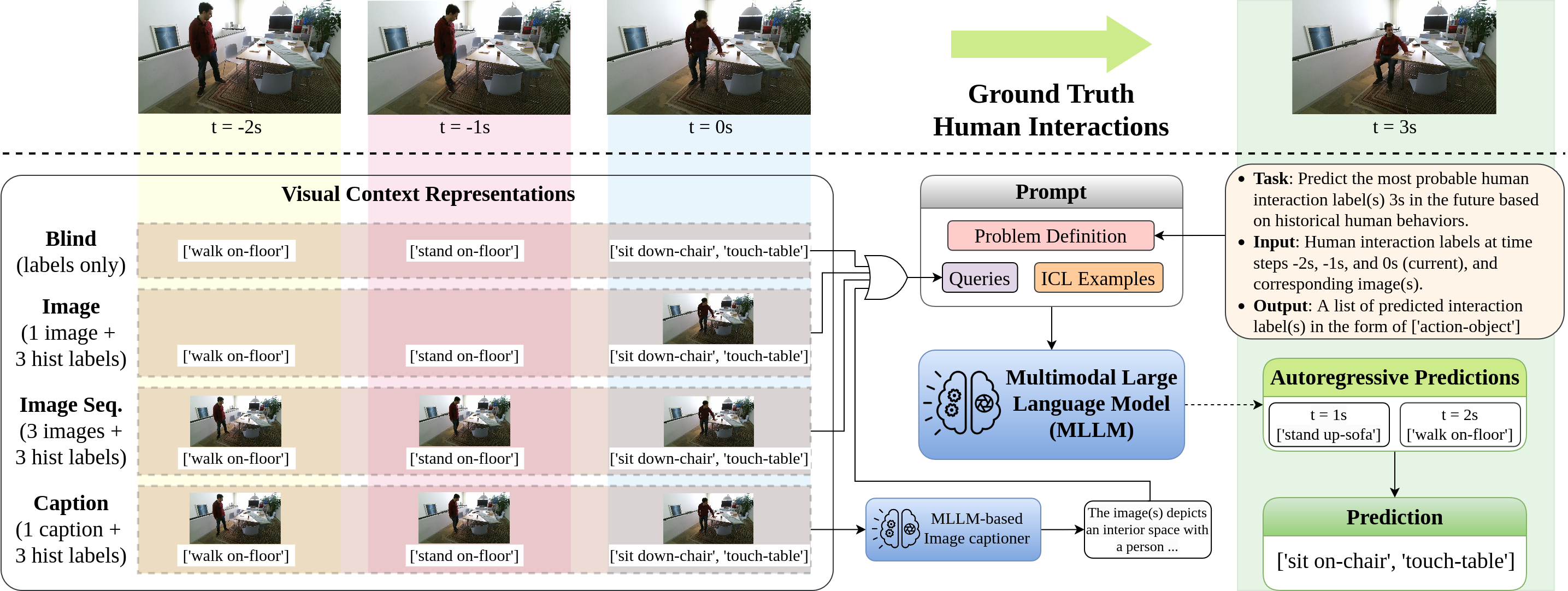}
    \caption{\textbf{System overview for human behavior prediction with scene context.} The top part depicts the \textbf{ground truth human interactions}. The prediction system is illustrated in the lower part, which consists of an \textbf{MLLM} that is instructed with a \textbf{prompt} including the task description to predict future interaction labels and a set of In-Context Learning (ICL) examples. We ablate the visual input with four types of \textbf{visual context representations}: blind (no visual input), image, image sequence, and image caption, varying from one to three past time steps. An \textbf{autoregressive prediction} step is added to improve the predictions, which predicts the intermediate action labels.\vspace{-1em}}
    \label{fig:sys}
\end{figure*}

\section{Related Work}
\label{sub:relatedWork}
Human behavior prediction has been a central focus in various fields, ranging from robotics to video analysis. Early approaches mainly relied on classical Deep Learning (DL)-based and data-driven techniques, leveraging models like Convolutional Neural Networks (CNNs) and Recurrent Neural Networks (RNNs) to process spatiotemporal patterns from visual input data, such as images, videos, or sensor signals, for activity recognition and trajectory prediction tasks \cite{rudenko2020human, kong2022human, alahi2016social, gupta2018social, sun2019relational, abu2018will}. Approaches like Social LSTM \cite{alahi2016social} and Social GAN \cite{gupta2018social} use RNNs and Generative Adversarial Networks (GANs), respectively, to predict pedestrian trajectories by learning interactions between individuals in shared spaces. 
% Sun \etal \cite{sun2019relational} further utilizes R-CNN to forecast multi-person actions from input videos.
However, these conventional methods often struggle with the growing prediction horizon and the complexity of human-object interactions \cite{rudenko2022atlas,kong2022human}. Furthermore, these data-driven approaches have difficulties in generalizing to unseen environments and human behaviors \cite{kujanpaa2024challenges}. The challenges can be tackled with the rapid development of more advanced methods, such as LLMs and MLLMs.

The rich embedded common-sense knowledge enables LLMs to achieve great performance as zero-shot reasoners \cite{kojima2022large}. 
Recent advances in LLM-based human behavior prediction include action anticipation, represented often as verb-noun pairs. The shift toward egocentric action anticipation, influenced by large-scale datasets, such as Epic-Kitchen \cite{damen2018scaling} and Ego4D \cite{grauman2022ego4d}, has opened new pathways for predicting human-object interactions and long-term action sequences from egocentric view.
Systems like PALM \cite{kim2024palm} demonstrate how LLMs can predict action sequences from visual inputs. Zhao \etal \cite{zhao2024antgpt} introduces AntGPT that leverages LLMs to model temporal dynamics and infer high-level goals from video embeddings. Moreover, multimodal approaches such as M-CAT \cite{beedu2024efficacy} combine textual and visual cues for action anticipation, enriching LLMs' predictions through contrastive learning. These LLM-based frameworks emphasize the utility of integrating visual context, past actions, and In-Context Learning (ICL), notably boosting the performance in predicting human behavior. 
Nevertheless, many of the LLM-driven action anticipation approaches rely on egocentric visual inputs, i.e., human actions from a first-person perspective. 
Research utilizing LLMs to predict human behaviors from a third-person perspective is relatively uncommon. For instance, Gorlo \etal \cite{gorlo2024long} leverages LLMs and scene graphs to predict human interaction sequences and trajectories in a synthetic dataset. Moreover, predicted human behaviors further benefit the downstream decision-making applications, enabling human awareness and achieving undisturbed task and motion planning \cite{liu2024towards, graule2024gg}.

In this paper, we propose a modular framework to benchmark pre-trained general-purpose MLLMs to predict future human behavior from a third-person robot perspective, using past human interaction labels and RGB images. Our framework does not require fine-tuning the MLLMs, can be applied on flexible prediction horizons and efficiently directed towards the action space in a specific domain with a small number of ICL examples. We systematically build the prediction pipeline and motivate the design choices for input representation, prompt composition, and task reasoning.

\section{Methodology}
\label{sec:method}

\subsection{Problem Definition}
We aim to reveal the capabilities of various MLLMs in predicting future human behaviors. The task is defined as predicting the interaction labels at a future time step based on the current visual context and the history of past interaction labels. 
An \textit{interaction label} refers to a symbolic \textit{``verb-noun"} pair, such as \textit{``touch-table"}, while a \textit{behavior} can be a list comprising multiple interaction labels, e.g., \textit{[``sit on-sofa", ``touch-table"]}.
Let \(L_t\) and \( V_t \) denote the interaction labels representing the human behavior and the visual scene context at time step \( t \), respectively. Using the MLLM-based prediction system \( f \), the objective is to predict the interaction labels \( \hat{L}_3 \) at a time step \( t = 3 \) seconds in the future, based on the current visual scene context \( V_0 \), the historical interaction labels \( L_{-2:0} \) from up to $2$ seconds in the past, and a set of ICL examples \( E^n \). The prediction is given by:

\begin{equation} \label{eq:pred}
    \hat{L}_3 = f\left(V_0, L_{-2:0}, E^n\right)
\end{equation}

where  \( E^n = \{e_1, e_2, \dots, e_n\} \). Each example \( e_i \) is a tuple of the label history from \( t = -2 \) to \( t = 0 \) and the ground truth future label at \( t = 3 \) which can be formulated as: 

\begin{equation} \label{eq:icl_exp}
    e_i = \left(L^{(i)}_{-2:0}, L^{(i)}_{_3}\right)
\end{equation}

The interaction label \(\hat{L}_3\) can also be predicted autoregressively with intermediate predictions \(\hat{L}_1\) and \(\hat{L}_2\) at the time steps \( t = 1 \) and \( t = 2 \), respectively. In this case, the prediction is reformulated as follows:

\begin{equation} \label{eq:pred_cot}
    \hat{L}_3 = f\left(V_0, L_{-2:0}, E^n, \hat{L}_1, \hat{L}_2 \right)
\end{equation}

A common problem with a short prediction horizon (e.g., $1$ or $2$ s) is that the activity is not likely to change in a short time, leading to a trivial prediction problem.
In order to meaningfully examine the prediction ability of the MLLMs, we therefore select a $3$-second prediction horizon. More details are explained in Sec. \ref{Sec:Datasets}.
An example of the problem definition in the prompt is shown on the right side of Fig.~\ref{fig:sys}.

\subsection{System Overview}
We propose a flexible and modular framework to benchmark MLLMs in predicting human behavior using 2D scene images and interaction histories. 
The system is designed to work with general-purpose MLLM backbones and handle the prediction task without task-specific model fine-tuning.
The key components of our system are illustrated in Fig.~\ref{fig:sys}. Firstly, there is the choice of a central MLLM module. The MLLM interprets the visual context to identify the human in the scene, ongoing activity and relevant objects. Secondly, there is the input structure which includes the history of past interactions, visual input, and captioning. Thirdly, there is the prompt design, which guides the reasoning process of MLLM in defining the specific prediction task and constraining the output format. The ICL technique is used to that end. Finally, there is the autoregressive prediction step which generates intermediate predictions.

\subsection{Multimodal Large Language Model (MLLM)}
\label{Sec:Models}
MLLMs typically consist of three main components \cite{caffagni_revolution_2024}: a language model backbone, a vision encoder, and a vision-to-language adapter (see Fig.~\ref{fig:ArchBehPred}). The language model backbone provides a foundation of general world knowledge and instruction-following capabilities. The visual encoder makes use of existing pre-trained models, with most MLLMs relying on language-supervised CLIP models \cite{radford_2021_PMLR}. The vision-to-language adapter ensures interoperability between the visual and language domains, and can be realized with different types of architectures, such as linear layers, Transformer-based modules, or cross-attention layers. 

Based on the state-of-the-art in the video analysis benchmarks, such as VideoMME \cite{fu2024video}, we adopt one proprietary and two open-source MLLM families, and two models from each family with different model sizes, namely GPT-4o and GPT-4o-mini \cite{openai_hello_2024}, Qwen2-VL-72B- and -7B-Instruct \cite{wang2024qwen2}, as well as LLaVA-NeXT-Video-34B- and -7B-DPO-hf \cite{li_llava-next-interleave_2024}.
While GPT-4o can process data with the most modalities, i.e., text, images, and audio, Qwen2-VL is designed to handle images, text, and bounding boxes as inputs, and can produce both text and bounding box outputs. Qwen2-VL's 72B model has been noted to outperform GPT-4o in certain metrics, especially in document understanding tasks. LLaVA-NeXT-Video is an extension of the LLaVA series \cite{liu2023visual}, focusing on video comprehension. It is built upon the Qwen2 language model and has demonstrated strong zero-shot performance in video tasks, even when trained only on image data.

\subsection{Visual Context Representations}
\label{sec:visualcontextrepresentations}

Visual context can be represented in several ways to enhance the accuracy of behavior predictions. We consider three visual representations relevant for the prediction task: a single \textbf{image}, a multi-frame \textbf{image sequence} capturing the motion over time, to provide a richer context for predicting future behavior, and a \textbf{scene caption} generated by the MLLM that describes the person, relevant objects and actions, and possible affordable actions. In addition, we evaluate a textual representation as a baseline (i.e., \textbf{blind}), where no visual context is provided, and predictions rely solely on historical interaction labels. The illustration of the visual context representations is shown in Fig.~\ref{fig:sys}.

\begin{figure*}[th!]
    \centering
    \includegraphics[width=0.75\linewidth]{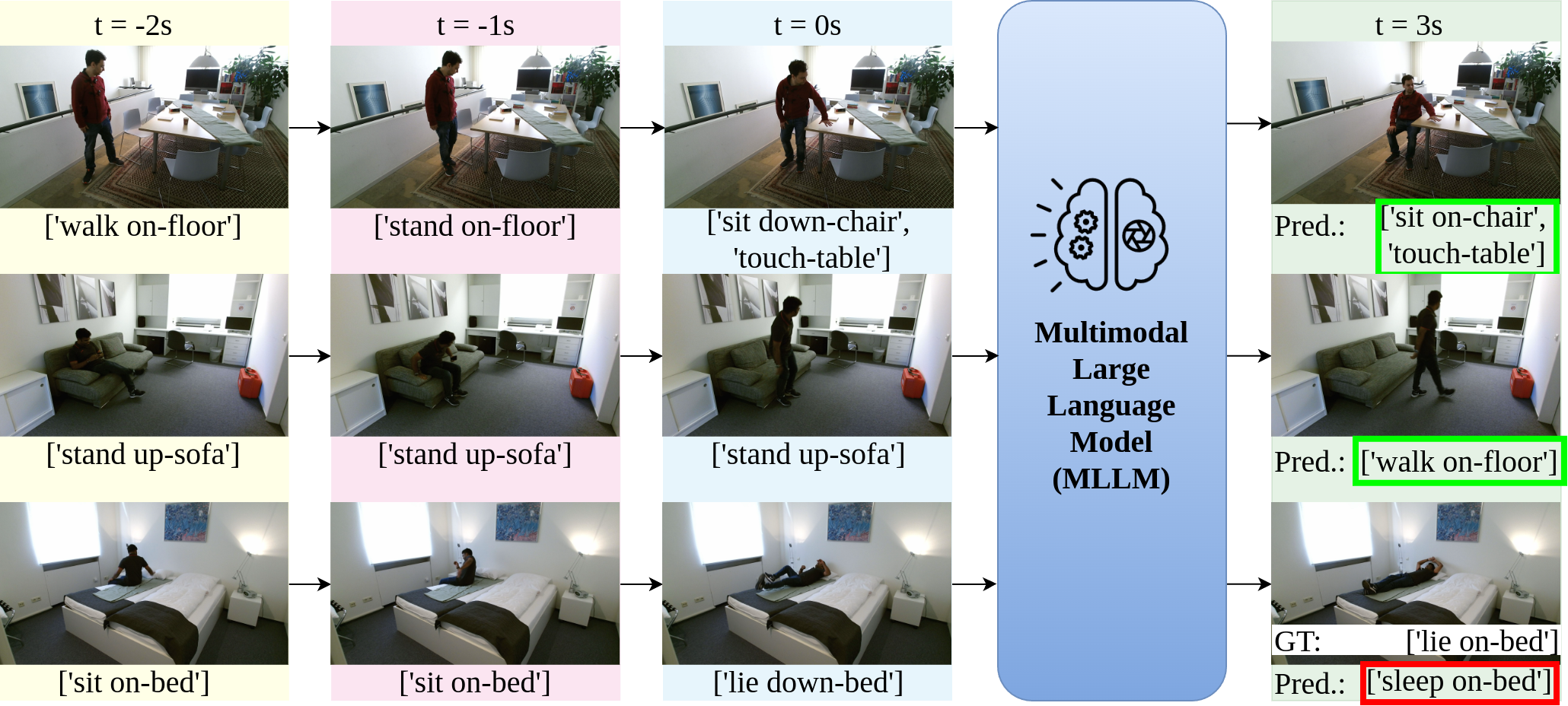}
    \caption{Examples of human behavior predictions. The last row depicts a false prediction that differs from the ground truth.\vspace{-1em}}
    \label{fig:qual_results}
\end{figure*}

\subsection{In-Context Learning (ICL)}
As LLMs are general-purpose models, they need to be directed to follow a specific task format. To this end, the concept of ICL is widely adopted, whereby the task is demonstrated by including examples of input-output pairs in the model prompt \cite{brown_language_2020}.
In our case, each visual ICL example \cite{alayrac_flamingo_2022} is a tuple with the interaction label history and the respective ground truth prediction, as shown in Fig.~\ref{fig:sys}. %The ICL examples aim to enhance the prediction accuracy in the following ways:
Using the ICL examples in the prediction task has several advantages: it demonstrates the desired output format, defines the set of common, domain-specific interaction labels, shows the common temporal patterns in the interactions, and thus helps improve the predictions. We motivate using ICL over fine-tuning because it can be applied at inference time, without the need of updating the model weights (i.e., a quite expensive procedure).

\subsection{Autoregressive Prediction}
\label{sub:intermedsteps}
Inspired by Chain-of-Thought (CoT) \cite{wei2022chain}, we try to guide the behavior prediction by adding intermediate prediction steps. Unlike CoT, these steps do not contain natural language reasoning but are intermediate predicted interactions that follow the same \textit{``action-object"} format. Given LLMs' autoregressive generation of output, the preceding steps are automatically incorporated into the context for the final 3-second prediction. In consequence, all predictions at timesteps 1, 2, and 3 seconds can be performed in a single response.

\section{Evaluation}
\label{sec:Experiments and Results}
We evaluate the prediction quality of our system and the impact of the possible design elements, outlined in Sec.~\ref{sec:method}.

\subsection{Experiments}
\label{Sec:ExperimentsBehavior}
We systematically analyze the impact of individual components on the prediction performance of the overall system. 
To this end, we propose the following experiments:

\boldparagraph{Visual Context Representations} We ablate the following four representations of the input sequence: blind, single image, image sequence, and caption, as illustrated in Fig.~\ref{fig:sys}.

\boldparagraph{Number of ICL Examples} The number varies from $1$ to $15$. 
Since GPT-4o can process a maximum of 50 images per request\footnote{\url{https://learn.microsoft.com/en-us/azure/ai-services/openai/quotas-limits}}, and the ICL examples have the same visual representation type as in the queries, with the sequence and caption types containing 3 images in each, the maximal number of ICL examples under the limit is 15, i.e., $3\,\text{images} \times (15\,\text{examples} + 1\,\text{query}) = 48$.

\boldparagraph{Autoregressive Prediction} We compare the performance with and without autoregressive prediction, i.e., intermediate predictions at timesteps 1, 2, and 3 seconds in the future.

The experiments benchmark GPT-4o (version 2024-05-13), GPT-4o-mini (version 2024-07-18), Qwen2-VL-72B-Instruct and -7B-Instruct (in the following denote as Qwen-72B and Qwen-7B), LLaVA-NeXT-Video-34B-DPO-hf and -7B-DPO-hf (denote as LLaVa-34B and LLaVa-7B). The GPT models are deployed with Azure OpenAI Service, while the Qwen and LLaVA models with two Nvidia H200 GPUs.

\subsection{Datasets}
\label{Sec:Datasets}
While recent studies focus on human behavior prediction from an egocentric view, for deployments on robotic systems and operation in human-cluttered environments, we choose human behavior datasets in a third-person view, i.e., PROX (Proximal Relationships with Object eXclusion) \cite{hassan_resolving_2019} and PROX-S datasets \cite{zhao_compositional_2022}, to build the evaluation sets and ICL examples. 
PROX is a dataset for 3D human pose estimation consistent with the 3D scene. It contains RGB image sequences of human motion and scene interaction in indoor environments from a third-person view and is sampled in 30 frames per second. The 12 scene environments cover bedrooms, living rooms, offices, and sitting booths. {PROX-S} extends the PROX dataset by adding semantic interaction labels on a frame-level. The interaction labels are formulated as action-object pairs, containing 17 different actions and 42 unique interactions, describing the common human-scene interactions.
There can be multiple interaction labels present for a frame if the human interacts with several objects, e.g. \textit{[``sit on-sofa", ``touch-table"]}. The MLLMs should predict the correct and the correct number of interaction labels.

We derive an evaluation dataset by combining the scene images from PROX with the semantic interaction labels from PROX-S. 
As shown in Fig.~\ref{fig:qual_results}, each data sequence is sampled with at least 4 frames across a 6-second time interval, i.e., 3 historical frames at time steps -2, -1, and 0 s, and a future frame at time steps 3 s. For autoregressive prediction, 2 more frames at time steps 1 and 2 s are sampled additionally. We sample the data sequences with a $1.5$-second time difference between the first frames. Using the test split of PROX-S, it yields an evaluation dataset of $329$ data sequences.
Aiming for a meaningful investigation of the prediction ability of different MLLMs with respect to \textbf{H1}, and targeting a practical short-term human activity anticipation for assistive robots, we set the horizon to 3 seconds, such that $47.4\%$ of the sequences in the evaluation dataset have different target and latest historical interaction labels (i.e., at time step 0s), and $25.8\%$ of the sequences have two or more interaction labels in the target frames, while with $1$ and $2s$ horizon, $74.8\%$ and $60.8\%$ of the data have identical target and past labels, respectively.
The ICL dataset is randomly sampled over the entire evaluation dataset according to an equal distribution.

\subsection{Metrics}
\label{subsec:metrics}
We evaluate the prediction performance with the following commonly-used metrics \cite{kong2022human}:

\boldparagraph{Cosine Similarity} It evaluates the cosine of the angle between the vector embeddings of two strings. It reveals the semantic similarity of two phrases and produce a value between $-1$ and $1$, where $1$ infers identical meaning. Let $\mathbf{A}$ and $\mathbf{B}$ denote the vector representations of the text strings, the cosine similarity is defined as:
\begin{equation}
    \text{Cosine Similarity} = \frac{\mathbf{A} \cdot \mathbf{B}}{\lVert\mathbf{A}\rVert \lVert\mathbf{B}\rVert}
\end{equation}

\boldparagraph{Edit Distance} The minimum number of operations (deletions, insertions, or substitutions) required to transform the predicted sequence $P$ into the ground truth sequence $G$. It reports the character-level similarity and is defined as:
\begin{equation}
    d_{\text{ed}}(P, G) = \min 
    \begin{cases} 
        d_{\text{ed}}(P_{1:m-1}, G_{1:n}) + 1 & \text{(del.)} \\
        d_{\text{ed}}(P_{1:m}, G_{1:n-1}) + 1 & \text{(ins.)} \\
        d_{\text{ed}}(P_{1:m-1}, G_{1:n-1}) + 1_{P_m \neq G_n} & \text{(sub.)}
    \end{cases}
\end{equation}
where $P_{1:m}$ and $G_{1:n}$ refer to the first $m$ and $n$ characters of the prediction $P$ and the ground truth $G$, respectively. The value of edit distance also varies between 0 and 1, where 0 indicates a full match.

\boldparagraph{Accuracy Score} It measures the accuracy-like score based on the overlap between prediction and ground truth interaction labels for the same future single frame, considering both exact and partial matches. It returns a value between $0$ and $1$, where $1$ means fully matched. It is computed as follows:
\begin{equation}
    \text{Accuracy} = \frac{2 \times |P \cap G|}{|P| + |G|}
\end{equation}
where $P$ and $G$ are the sets of predicted and ground-truth interaction labels, respectively. 
Achieving high accuracy when considering multiple human activity prediction is quite challenging: recent state-of-the-art baselines with ego-centric view can reach an accuracy on average of $0.6$ \cite{zhao2024antgpt}.

Computational efficiency and a more dedicated analysis considering different distillation techniques are out of the scope of the present work and left for future research.

\begin{figure*}
    \centering
    \includegraphics[width=0.89\linewidth]{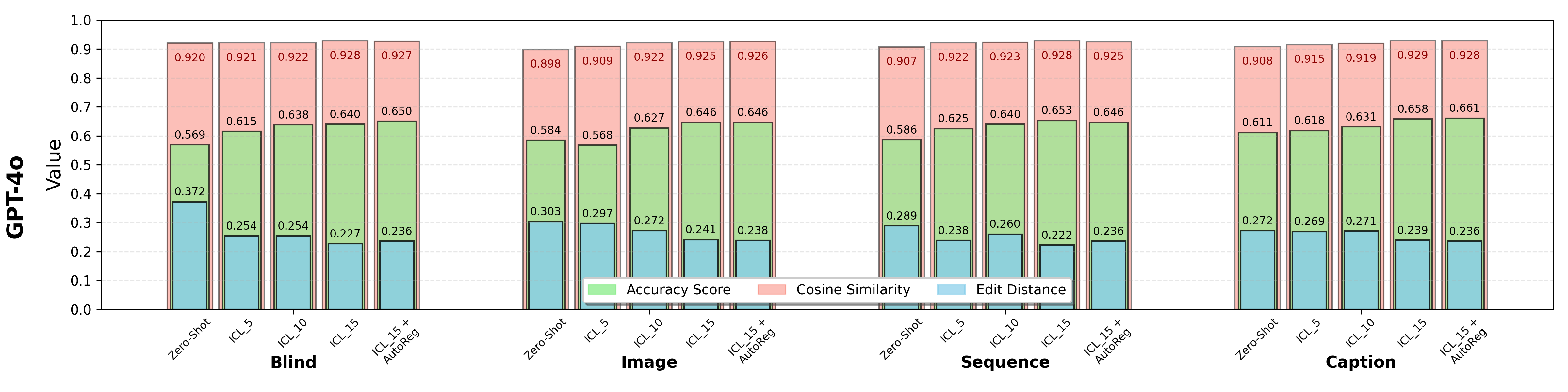}
    \includegraphics[width=0.89\linewidth]{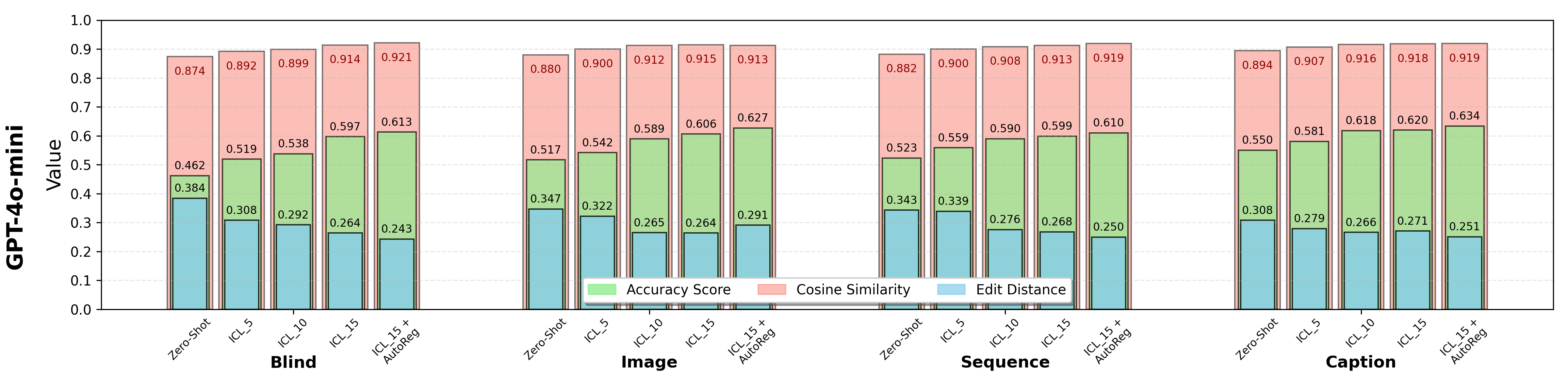}
    \includegraphics[width=0.89\linewidth]{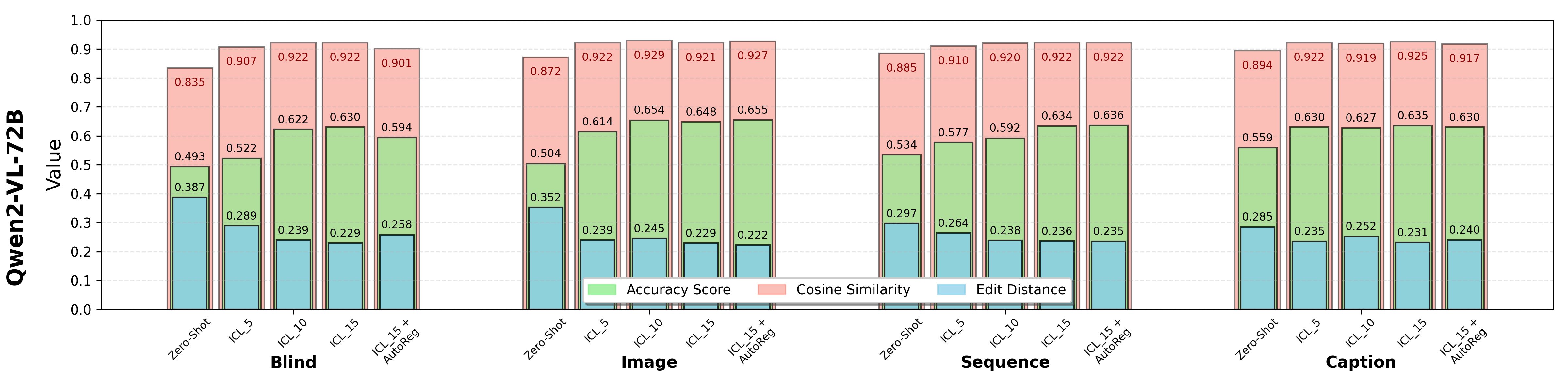}
    \includegraphics[width=0.89\linewidth]{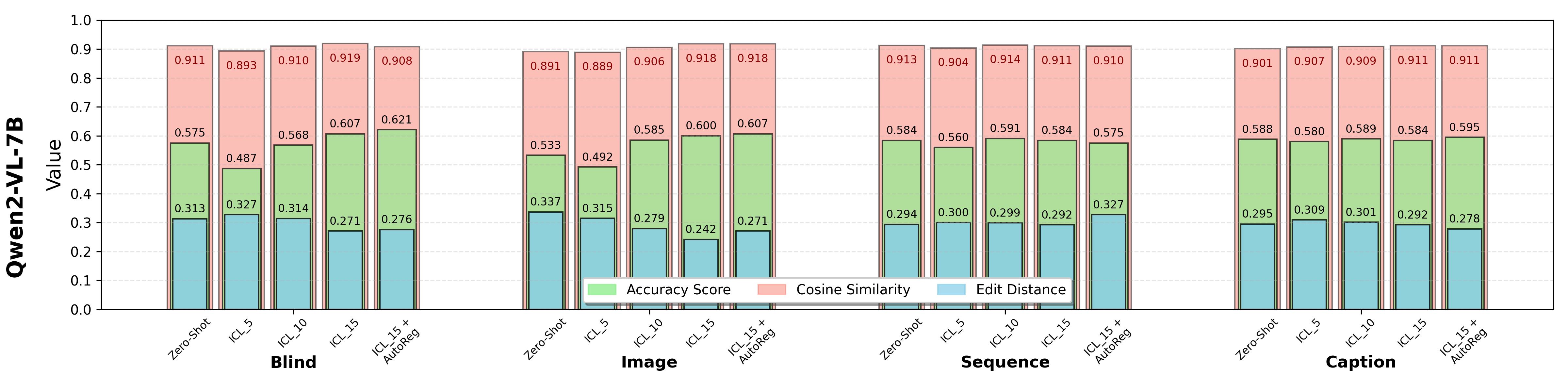}
    \includegraphics[width=0.89\linewidth]{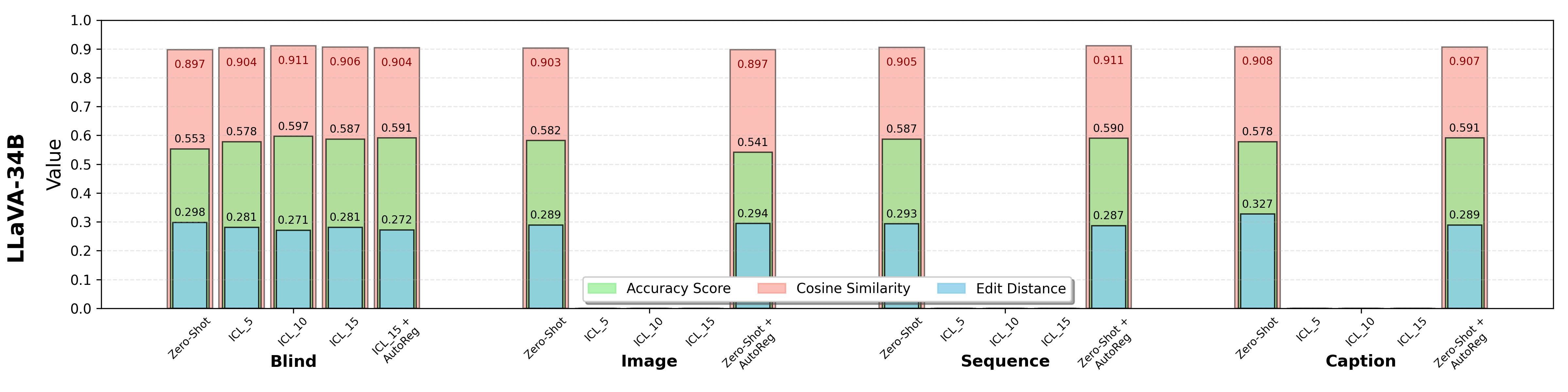}
    \includegraphics[width=0.89\linewidth]{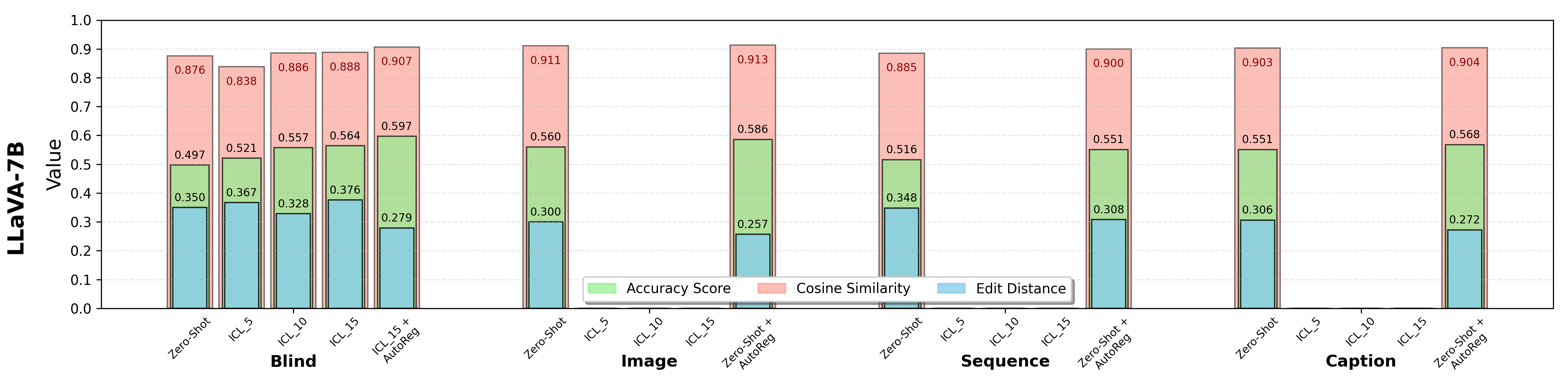}
    \caption{Quantitative comparison of all models conditioned on input representations, number of ICL examples, and autoregressive prediction.}
    \label{fig:results_main}
\end{figure*}

\section{Results}
\label{Sec:EvalBehPred}

The comprehensive quantitative evaluation results, which include all MLLMs and visual representation types conditioned on the number of ICL examples are presented in Fig.~\ref{fig:results_main}.

\begin{table}[t]
    \centering
    \renewcommand{\arraystretch}{0.95}
    \setlength{\tabcolsep}{2.2pt}
    \begin{tabular}{c l *{3}{c}}
        \toprule
        Visual Repr. & MLLM & Accuracy $\uparrow$ & Cos. Sim. $\uparrow$ & Edit Dist. $\downarrow$ \\
        \midrule
        \multirow{6}{*}{Blind} & GPT-4o & 0.569 & \textbf{0.920} & 0.372 \\
         & GPT-4o-mini & 0.462 & 0.874 & 0.384 \\
         & Qwen-72B & 0.493 & 0.835 & 0.387\\
         & Qwen-7B & \textbf{0.575} &	0.911 & 0.313 \\
         & LLaVA-34B & 0.553	& 0.897	& \textbf{0.298} \\
         & LLaVA-7B & 0.497& 0.876 & 0.350 \\
        \midrule
        \multirow{6}{*}{Image} & GPT-4o & \textbf{0.584} & 0.898 & 0.303 \\
         & GPT-4o-mini & 0.517 & 0.880 & 0.347 \\
         & Qwen-72B & 0.504 & 0.872 & 0.352 \\
         & Qwen-7B & 0.533 &	0.891 &	0.337 \\
         & LLaVA-34B & 0.582	& 0.903	& \textbf{0.289} \\
         & LLaVA-7B & 0.560 & \textbf{0.911} & 0.300 \\
        \midrule
        \multirow{6}{*}{Sequence} & GPT-4o & 0.586 & 0.907 & \textbf{0.289}\\
         & GPT-4o-mini & 0.523 & 0.882 & 0.343 \\
         & Qwen-72B & 0.534&	0.885&	0.297 \\
         & Qwen-7B & 0.584&	\textbf{0.913} & 0.294 \\
         & LLaVA-34B & \textbf{0.587}	& 0.905	& 0.293 \\
         & LLaVA-7B & 0.516&	0.885&	0.348 \\
        \midrule
        \multirow{6}{*}{Caption} & GPT-4o & \textbf{0.611} & \textbf{0.908} & \textbf{0.272} \\
         & GPT-4o-mini & 0.550 & 0.894 & 0.308 \\
         & Qwen-72B & 0.559 & 0.894 & 0.285 \\
         & Qwen-7B & 0.588 &	0.901 & 0.295 \\
         & LLaVA-34B & 0.578	& \textbf{0.908} & 0.327 \\
         & LLaVA-7B & 0.551 & 0.903 & 0.306 \\
        \bottomrule
    \end{tabular}
    \caption{Results of Zero-Shot Prediction. Higher values of accuracy and cosine similarity and lower value of edit distance indicate better performance. Bold numbers indicate the best results in each visual representation.\vspace{-1em}}
    \label{tab:zeroshot}
\end{table}

\subsection{Visual Context Representations}
In this section, we compare the zero-shot prediction performance of the MLLMs across different visual representation types.
Extracting the first columns from each cluster in Fig.~\ref{fig:results_main} results in Table~\ref{tab:zeroshot}.
Despite being extensively trained with visual datasets, Qwen-7B achieves the highest accuracy at $0.575$ in blind representation, slightly surpassing GPT-4o at $0.569$, while all other MLLMs perform the worst compared to other representations. Qwen-72B reports the lowest value of cosine similarity and highest edit distance, and GPT-4o-mini returns the worst accuracy compared to the other representation types. This suggests that the absence of visual input limits the performance, highlighting the importance of visual information in context-aware human behavior prediction.

With the image representation, the models perform slightly better than with the blind representation. Among them, GPT-4o achieved the highest accuracy at $0.584$, and GPT-4o-mini gains the most improvement in accuracy by $11.9\%$ ($0.517$ vs. $0.462$). However, the accuracy of Qwen-7B is even lower than the blind approach by $0.042$ ($0.533$ vs. $0.575$). No major changes are shown in terms of cosine similarity or edit distance. This possibly infers that compared to textual interaction histories, adding a single image alone may introduce ambiguity or lack sufficient context of human behavior or the environment.

The results with sequence representation are generally better than in both single-image and blind settings. The improvement is more pronounced for Qwen-7B, where the accuracy score increases by $9.6\%$ ($0.584$ vs. $0.533$ in image representation), and it also achieves the best cosine similarity. For the other MLLMs, the performance gain is moderate but consistent.
It means that alongside the textual labels and a single image, having multiple images in a continuous sequence provides a richer temporal context. Since human behaviors are usually dynamic, a sequence of images helps the models to capture intent, motion, and cause-effect relationships better than a single static image.

Caption and sequence representations are both designed to capture the temporal context in human behavior prediction tasks. However, adding a textual scene caption provides additional descriptive information that may enhance scene understanding of the MLLMs, and can potentially help reduce ambiguity in behavior prediction. Both representations show very similar performance patterns, where caption consistently outperforms sequence setting by circa $2\% - 3\%$ in accuracy. Among them, GPT-4o achieves the highest accuracy at $0.611$, which is also the best overall result in zero-shot prediction. However, the accuracy and edit distance of LLaVA-34 are worse in caption than in sequence representation, where adding further captions provides only redundant information that the model has already extracted from the visual sequences themselves.
On the other hand, only minor variations are observed in cosine similarity and edit distance.

In summary, incorporating visual inputs generally enhances model performance in human behavior prediction tasks. Thus, we confirm \textbf{H2}. Notably, the addition of textual captions to image sequences provides significant improvement for GPT-4o, but only marginally for the other models.

\subsection{Number of In-Context Learning Examples}
Observing the results from the changing number of ICL examples across the models in Fig.~\ref{fig:results_main}, we observe a general trend that the performance improves with the growing number of ICL examples, providing evidence in favor of \textbf{H3}. All models show a consistent improvement in all three metrics as the number of ICL examples increases from $0$ (i.e., zero-shot) to $15$, except for Qwen-7B, whose accuracy score considerably drops when the number of ICL examples grows from $0$ to $5$, especially in blind and image settings. Furthermore, both Qwen models encounter a performance saturation in all three visual-based representations, especially with more than $10$ examples.
This is possibly related to the inconsistent quality of the examples since they are randomly sampled over the entire evaluation dataset. Another possible reason can be potential overfitting with small example sets, causing it to focus too heavily on specific features from those few examples rather than generalizing appropriately. 

Unfortunately, the LLaVA models seem to be unsuitable for ICL. This is potentially caused by its prompt template that always first renders all images before processing the text \footnote{\url{https://cap-mllm.github.io/}}. The missing tokenize-in-place ability leads to the mismatching of example and query images, therefore consistently resulting in failures and generating corrupted output tokens. Hence, we do not report the ICL results of the LLaVA models in Fig.~\ref{fig:results_main}. Moreover, we also observed that LLaVA models perform poorly in generating structured output, making it difficult to extract the predicted interaction labels.

\subsection{Autoregressive Prediction}
By predicting the intermediate interaction labels at timesteps of $1s$ and $2s$, the autoregressive prediction shows minimal improvement over the accuracy score and edit distance with $15$ ICL examples of GPT and Qwen models, and zero-shot prediction results of LLaVA models, and has no major impact on the cosine similarity, as depicted in the last columns of each cluster in Fig.~\ref{fig:results_main}. On average, autoregressive prediction improves the accuracy score for each MLLM by circa $1\%$. This indicates that autoregressive prediction can smooth out the final outcome, but only acts as a minor refinement step rather than a major performance booster. At this point, it is not possible to conclude in favor of \textbf{H4}.

\subsection{Summary and qualitative results}
Our evaluation shows that many configurations can deliver competitive values of cosine similarity (over $0.9$) and edit distance (close to $0.2$). Achieving high accuracy is the most challenging task. The combination of GPT-4o with caption representation, $15$ ICL examples, and autoregressive prediction obtains the highest accuracy at $66.1\%$.
In Fig.~\ref{fig:qual_results}, we provide several examples to showcase some of the challenging aspects of the behavior prediction problem: third-person view from the robot in cluttered environments with challenging lighting, partial occlusions, limited visibility of objects, and potentially multiple ground truth labels. Overall, we find evidence in favor of \textbf{H1}, concluding that MLLMs can successfully predict non-trivial sequences with potentially multiple ground truth labels.

\section{Limitations}
\label{Sub:Disc-limitations}

\subsection{Limitations of In-Context Learning}
While ICL improves the performance, it faces scaling limitations, as the performance gains plateau after $10$ examples for most cases. 
However, computational costs increase with more examples, posing challenges for real-time applications. Additionally, the effectiveness of ICL depends on example quality and diversity, and also vaires across different model families. Vision-based MLLMs such as Qwen models show marginal improvement with ICL comparing to general-purpose MLLMs such as GPT models. Thus, optimizing ICL requires balancing example selection, context constraints, and computational efficiency to maximize the performance.

\subsection{Limitations of MLLMs in visual and spatial reasoning}
The experiment results show that the MLLMs heavily depend on additional context (i.e., ICL examples) to achieve higher accuracy, with zero-shot performance being notably weaker. This indicates the limitations of pre-trained MLLMs in visual and spatial reasoning. While the GPT models benefit from richer multimodal input, the Qwen models show minimal difference between textual (i.e., blind) and visual representations, suggesting that they struggle to extract meaningful spatial information directly from visual inputs without textual guidance. This is further evidenced by sequence and caption representations producing similar results in the Qwen models, where all metrics level out, demonstrating a clear ceiling effect in spatial reasoning capabilities, and indicating inefficient utilization of temporal-spatial information. These findings highlight the challenges of grounding spatial understanding in MLLMs across architectures.

\section{Conclusion}
\label{Sub:main-finings}
In this paper, we presented a modular and generic framework for human behavior prediction based on Multimodal Large Language Models (MLLMs). We performed a systematic evaluation in terms of the quality of the generated predictions by varying the type of input context, number of In-Context Learning (ICL) examples, and the addition of autoregressive prediction. The findings suggest that the best-performing model configuration reaches $92.8\%$ semantic similarity and $66.1\%$ exact label accuracy in predicting potentially multiple human behaviors in the target frame.

Our evaluation supports the hypotheses \textbf{H1}, \textbf{H2}, and \textbf{H3} presented in Sec.~\ref{sec:intro}. Firstly, many configurations can deliver competitive results without additional fine-tuning for correctly predicting non-trivial sequences of human activities. Furthermore, incorporating additional visual context and more ICL examples yields the best performance.

In future work, we aim to extend this evaluation to real-world experiments, consider other input modalities (e.g., audio, depth), and analyze computational efficiency.

% \clearpage
% \footnotesize
\bibliographystyle{IEEEtran}
\bibliography{main}

\def\authornoop#1{}
\begin{thebibliography}{10}
\providecommand{\url}[1]{#1}
\csname url@samestyle\endcsname
\providecommand{\newblock}{\relax}
\providecommand{\bibinfo}[2]{#2}
\providecommand{\BIBentrySTDinterwordspacing}{\spaceskip=0pt\relax}
\providecommand{\BIBentryALTinterwordstretchfactor}{4}
\providecommand{\BIBentryALTinterwordspacing}{\spaceskip=\fontdimen2\font plus
\BIBentryALTinterwordstretchfactor\fontdimen3\font minus \fontdimen4\font\relax}
\providecommand{\BIBforeignlanguage}[2]{{%
\expandafter\ifx\csname l@#1\endcsname\relax
\typeout{** WARNING: IEEEtran.bst: No hyphenation pattern has been}%
\typeout{** loaded for the language `#1'. Using the pattern for}%
\typeout{** the default language instead.}%
\else
\language=\csname l@#1\endcsname
\fi
#2}}
\providecommand{\BIBdecl}{\relax}
\BIBdecl

\bibitem{rudenko2020human}
A.~Rudenko, L.~Palmieri, M.~Herman, K.~M. Kitani, D.~M. Gavrila, and K.~O. Arras, ``Human motion trajectory prediction: A survey,'' \emph{The International Journal of Robotics Research}, vol.~39, no.~8, 2020.

\bibitem{zhang2023pedestrian}
C.~Zhang and C.~Berger, ``Pedestrian behavior prediction using deep learning methods for urban scenarios: A review,'' \emph{IEEE Transactions on Intelligent Transportation Systems}, vol.~24, no.~10, 2023.

\bibitem{jahanmahin2022human}
R.~Jahanmahin, S.~Masoud, J.~Rickli, and A.~Djuric, ``Human-robot interactions in manufacturing: A survey of human behavior modeling,'' \emph{Robotics and Computer-Integrated Manufacturing}, vol.~78, 2022.

\bibitem{cao_long-term_2020}
Z.~Cao, H.~Gao, K.~Mangalam, Q.-Z. Cai, M.~Vo, and J.~Malik, ``Long-term {Human} {Motion} {Prediction} with {Scene} {Context},'' Jul. 2020, arXiv:2007.03672 [cs].

\bibitem{stefanini2024ral}
E.~Stefanini, L.~Palmieri, A.~Rudenko, T.~Hielscher, T.~Linder, and L.~Pallottino, ``Efficient context-aware model predictive control for human-aware navigation,'' in \emph{{IEEE} Robotics and Automation Letters}, 2024.

\bibitem{lasota2019robust}
P.~A. Lasota, ``Robust human motion prediction for safe and efficient human-robot interaction,'' Ph.D. dissertation, Massachusetts Institute of Technology, 2019.

\bibitem{kruse2013human}
T.~Kruse, A.~K. Pandey, R.~Alami, and A.~Kirsch, ``Human-aware robot navigation: A survey,'' \emph{Robotics and Autonomous Systems}, vol.~61, no.~12, 2013.

\bibitem{li2021toward}
S.~Li, P.~Zheng, J.~Fan, and L.~Wang, ``Toward proactive human--robot collaborative assembly: A multimodal transfer-learning-enabled action prediction approach,'' \emph{IEEE Transactions on Industrial Electronics}, vol.~69, no.~8, 2021.

\bibitem{liu2024delta}
Y.~Liu, L.~Palmieri, S.~Koch, I.~Georgievski, and M.~Aiello, ``Delta: Decomposed efficient long-term robot task planning using large language models,'' \emph{arXiv preprint arXiv:2404.03275}, 2024.

\bibitem{zhao2024wildhallucinations}
W.~Zhao \emph{et~al.}, ``Wildhallucinations: Evaluating long-form factuality in llms with real-world entity queries,'' \emph{arXiv:2407.17468}, 2024.

\bibitem{chen2024motionllm}
L.-H. Chen \emph{et~al.}, ``Motionllm: Understanding human behaviors from human motions and videos,'' \emph{arXiv preprint arXiv:2405.20340}, 2024.

\bibitem{hassan_resolving_2019}
M.~Hassan, V.~Choutas, D.~Tzionas, and M.~Black, ``Resolving {3D} {Human} {Pose} {Ambiguities} {With} {3D} {Scene} {Constraints},'' in \emph{2019 {IEEE}/{CVF} {International} {Conf.} on {Computer} {Vision} ({ICCV})}, Oct. 2019.

\bibitem{zhao_compositional_2022}
K.~Zhao, S.~Wang, Y.~Zhang, T.~Beeler, and S.~Tang, ``\BIBforeignlanguage{en}{Compositional {Human}-{Scene} {Interaction} {Synthesis} with {Semantic} {Control}},'' in \emph{\BIBforeignlanguage{en}{Computer {Vision} – {ECCV} 2022}}, ser. Lecture {Notes} in {Computer} {Science}, S.~Avidan, G.~Brostow, M.~Cissé, G.~M. Farinella, and T.~Hassner, Eds.\hskip 1em plus 0.5em minus 0.4em\relax Cham: Springer Nature Switzerland, 2022, pp. 311--327.

\bibitem{kong2022human}
Y.~Kong and Y.~Fu, ``Human action recognition and prediction: A survey,'' \emph{Int. Journal of Computer Vision}, vol. 130, no.~5, 2022.

\bibitem{alahi2016social}
A.~Alahi, K.~Goel, V.~Ramanathan, A.~Robicquet, L.~Fei-Fei, and S.~Savarese, ``Social lstm: Human trajectory prediction in crowded spaces,'' in \emph{Proc. of the IEEE conf. on computer vision and pattern recognition}, 2016.

\bibitem{gupta2018social}
A.~Gupta, J.~Johnson, L.~Fei-Fei, S.~Savarese, and A.~Alahi, ``Social gan: Socially acceptable trajectories with generative adversarial networks,'' in \emph{IEEE Conf. on computer vision and pattern recognition}, 2018.

\bibitem{sun2019relational}
C.~Sun, A.~Shrivastava, C.~Vondrick, R.~Sukthankar, K.~Murphy, and C.~Schmid, ``Relational action forecasting,'' in \emph{Proceedings of the IEEE/CVF Conf. on Computer Vision and Pattern Recognition}, 2019.

\bibitem{abu2018will}
Y.~Abu~Farha, A.~Richard, and J.~Gall, ``When will you do what?-anticipating temporal occurrences of activities,'' in \emph{Proceedings of the IEEE conference on computer vision and pattern recognition}, 2018.

\bibitem{rudenko2022atlas}
A.~Rudenko, L.~Palmieri, W.~Huang, A.~J. Lilienthal, and K.~O. Arras, ``The atlas benchmark: An automated evaluation framework for human motion prediction,'' in \emph{2022 IEEE International Conference on Robot and Human Interactive Communication (RO-MAN)}.\hskip 1em plus 0.5em minus 0.4em\relax IEEE, 2022.

\bibitem{kujanpaa2024challenges}
K.~Kujanp{\"a}{\"a} \emph{et~al.}, ``Challenges of data-driven simulation of diverse and consistent human driving behaviors,'' \emph{arXiv:2401.03236}, 2024.

\bibitem{kojima2022large}
T.~Kojima, S.~S. Gu, M.~Reid, Y.~Matsuo, and Y.~Iwasawa, ``Large language models are zero-shot reasoners,'' \emph{Advances in neural information processing systems}, vol.~35, 2022.

\bibitem{damen2018scaling}
D.~Damen \emph{et~al.}, ``Scaling egocentric vision: The epic-kitchens dataset,'' in \emph{Proc. of the European Conf. on computer vision (ECCV)}, 2018.

\bibitem{grauman2022ego4d}
K.~Grauman \emph{et~al.}, ``Ego4d: Around the world in 3,000 hours of egocentric video,'' in \emph{Proceedings of the IEEE/CVF Conference on Computer Vision and Pattern Recognition}, 2022.

\bibitem{kim2024palm}
S.~Kim, D.~Huang, Y.~Xian, O.~Hilliges, L.~Van~Gool, and X.~Wang, ``Palm: Predicting actions through language models,'' in \emph{European Conference on Computer Vision}.\hskip 1em plus 0.5em minus 0.4em\relax Springer, 2024.

\bibitem{zhao2024antgpt}
Q.~Zhao \emph{et~al.}, ``Antgpt: Can large language models help long-term action anticipation from videos?'' in \emph{ICLR}, 2024.

\bibitem{beedu2024efficacy}
A.~Beedu, K.~Samel, and I.~Essa, ``On the efficacy of text-based input modalities for action anticipation,'' \emph{arXiv:2401.12972}, 2024.

\bibitem{gorlo2024long}
N.~Gorlo, L.~Schmid, and L.~Carlone, ``Long-term human trajectory prediction using 3d dynamic scene graphs,'' \emph{arXiv preprint arXiv:2405.00552}, 2024.

\bibitem{liu2024towards}
Y.~Liu, L.~Palmieri, S.~Koch, I.~Georgievski, and M.~Aiello, ``Towards human awareness in robot task planning with large language models,'' \emph{arXiv preprint arXiv:2404.11267}, 2024.

\bibitem{graule2024gg}
M.~A. Graule and V.~Isler, ``Gg-llm: Geometrically grounding large language models for zero-shot human activity forecasting in human-aware task planning,'' in \emph{2024 IEEE International Conference on Robotics and Automation (ICRA)}.\hskip 1em plus 0.5em minus 0.4em\relax IEEE, 2024.

\bibitem{caffagni_revolution_2024}
D.~Caffagni \emph{et~al.}, ``The revolution of multimodal large language models: A survey,'' in \emph{Findings of the Association for Computational Linguistics: ACL 2024}, 2024.

\bibitem{radford_2021_PMLR}
A.~Radford \emph{et~al.}, ``Learning transferable visual models from natural language supervision,'' in \emph{Int. Conf. on machine learning}.\hskip 1em plus 0.5em minus 0.4em\relax PMLR, 2021.

\bibitem{fu2024video}
C.~Fu \emph{et~al.}, ``Video-mme: The first-ever comprehensive evaluation benchmark of multi-modal llms in video analysis,'' \emph{arXiv preprint arXiv:2405.21075}, 2024.

\bibitem{openai_hello_2024}
\BIBentryALTinterwordspacing
OpenAI, ``\BIBforeignlanguage{en-US}{Hello {GPT}-4o},'' May 2024. [Online]. Available: \url{https://openai.com/index/hello-gpt-4o/}
\BIBentrySTDinterwordspacing

\bibitem{wang2024qwen2}
P.~Wang \emph{et~al.}, ``Qwen2-vl: Enhancing vision-language model's perception of the world at any resolution,'' \emph{arXiv:2409.12191}, 2024.

\bibitem{li_llava-next-interleave_2024}
\BIBentryALTinterwordspacing
F.~Li \emph{et~al.}, ``{LLaVA}-{NeXT}-{Interleave}: {Tackling} {Multi}-image, {Video}, and {3D} in {Large} {Multimodal} {Models},'' Jul. 2024, arXiv:2407.07895 [cs]. [Online]. Available: \url{http://arxiv.org/abs/2407.07895}
\BIBentrySTDinterwordspacing

\bibitem{liu2023visual}
H.~Liu, C.~Li, Q.~Wu, and Y.~J. Lee, ``Visual instruction tuning,'' \emph{Advances in neural information processing systems}, vol.~36, 2023.

\bibitem{brown_language_2020}
T.~Brown \emph{et~al.}, ``Language {Models} are {Few}-{Shot} {Learners},'' \emph{Advances in Neural Information Processing Systems}, vol.~33, 2020.

\bibitem{alayrac_flamingo_2022}
J.-B. Alayrac \emph{et~al.}, ``Flamingo: a {Visual} {Language} {Model} for {Few}-{Shot} {Learning},'' \emph{Advances in Neural Information Processing Systems}, vol.~35, pp. 23\,716--23\,736, 2022.

\bibitem{wei2022chain}
J.~Wei \emph{et~al.}, ``Chain-of-thought prompting elicits reasoning in large language models,'' \emph{Advances in neural information processing systems}, vol.~35, pp. 24\,824--24\,837, 2022.

\end{thebibliography}

\end{document}